\documentclass{article}

\usepackage[english]{babel}
\usepackage{authblk}
\usepackage[letterpaper,top=2cm,bottom=2cm,left=3cm,right=3cm,marginparwidth=1.75cm]{geometry}

\usepackage{amsmath}
\usepackage{graphicx}
\usepackage[colorlinks=true, allcolors=blue]{hyperref}

\usepackage{apacite}

\usepackage{xcolor}
\newcounter{aqctr}
\newenvironment{author-query}
{\refstepcounter{aqctr}\par\vspace{\baselineskip}\noindent
\color{red}\textbf{Author Query/Comment AQ \arabic{aqctr}.}}
{\par\vspace{\baselineskip}\normalcolor}

\title{Cost-efficient generative AI summarization for scalable automated essay scoring in educational assessment}
\author[1]{Haowei Hua}
\affil[1]{Princeton University, Princeton, NJ, USA}
\affil[ ]{\texttt{jack.hua@princeton.edu}}
\begin{document}
\maketitle

\begin{abstract}
Automated essay scoring (AES) has become an important tool in educational assessment by enabling scalable evaluation of student writing and timely formative feedback. However, many transformer-based embedding models (e.g., BERT, RoBERTa, DeBERTa) are constrained by strict input-length limits, making it difficult to process long-form essays without truncation and potential loss of educationally relevant information. This study proposes a generative AI-assisted summarization framework to address this limitation while preserving scoring reliability. Using the ASAP 2.0 dataset, we generate controlled-length summaries of student essays with three GPT-5 model variants (GPT-5, GPT-5 mini, GPT-5 nano) and use these summaries as inputs for downstream scoring models. To retain important linguistic signals, we integrate handcrafted features from the original essays, forming a hybrid representation of writing. The framework is evaluated across downstream scoring performance, summarization quality, and model-cost considerations. Scoring performance is measured using quadratic weighted kappa (QWK), while summarization quality is assessed through lexical overlap, semantic similarity, information retention, and redundancy. Results show that GPT-5 mini achieves the highest agreement with human raters, while GPT-5 provides the strongest summarization quality. Summarization performance declines for higher-scoring essays, suggesting that more complex writing is harder to compress without information loss. These findings highlight trade-offs among model capacity, summary fidelity, computational cost, and construct preservation. The study should be interpreted as an initial controlled evaluation of GPT-based summarization for AES rather than a complete benchmark against all long-context and direct-scoring alternatives. Overall, the study demonstrates that generative AI summarization can support reliable and scalable assessment of writing ability in educational contexts, while also identifying the baseline and ablation experiments needed for stronger generalization claims.
\end{abstract}

\section{Introduction}

Automated essay scoring (AES) refers to the use of computational techniques to evaluate written responses in ways that approximate or support human judgment. The concept dates back to early computational assessment research, where pioneering systems demonstrated the feasibility of scoring essays using statistical properties of text rather than direct human evaluation \cite{page1966imminence}. These early efforts laid the groundwork for understanding how measurable linguistic patterns could approximate aspects of writing quality. The motivation for AES arises from a fundamental challenge in education: writing assessment is labor-intensive, time-consuming, and difficult to scale consistently across large student populations. Human grading requires trained raters, standardized rubrics, and extensive quality control procedures, all of which limit the speed and scalability of assessment. As educational systems expand and digital learning environments proliferate, the demand for efficient and reliable assessment tools has increased substantially. AES systems aim to address this need by providing rapid scoring, consistent evaluation criteria, and scalable feedback mechanisms that support both classroom instruction and large-scale standardized testing \cite{dikli2006overview,attali2006automated,williamson2012framework}. By reducing grading latency and improving consistency across raters, AES technologies contribute to more timely educational feedback and more efficient resource allocation in assessment processes.

In educational contexts, AES has been widely adopted for formative assessment, placement testing, and high-stakes examinations where large volumes of written responses must be evaluated under strict time constraints. Beyond efficiency, automated scoring can reduce grading delays, standardize evaluation practices, and support personalized learning environments by providing immediate diagnostic feedback to learners. Such rapid feedback cycles are especially valuable in writing-intensive curricula, where iterative revision plays a central role in skill development. However, writing quality is inherently multidimensional, involving grammar, vocabulary, coherence, organization, argumentation, rhetorical effectiveness, and audience awareness. Capturing these diverse dimensions computationally is a complex challenge, requiring both linguistic sensitivity and alignment with educational constructs of writing proficiency. Consequently, AES research has evolved into an interdisciplinary field spanning educational measurement, computational linguistics, artificial intelligence, and cognitive psychology. A central concern in this domain is ensuring that automated scores align with the intended construct of writing proficiency while maintaining fairness, validity, reliability, and interpretability across diverse student populations \cite{williamson2012framework,lottridge2023psychometric,madnani2018automated,shermis2013handbook}. This emphasis on construct validity distinguishes educational assessment applications from purely technical text classification tasks. Early AES systems relied heavily on handcrafted linguistic features and statistical modeling approaches. Operational systems such as e-rater extracted measurable indicators related to grammar usage, lexical diversity, syntactic complexity, discourse organization, and stylistic patterns, and used these features to train regression or classification models that predicted holistic writing scores \cite{attali2006automated,burstein2004automated}. These feature-based systems demonstrated that computational models could approximate human scoring performance under controlled conditions, particularly when large annotated datasets were available. However, their performance was often constrained by limited semantic understanding and heavy reliance on manually designed rules that could not fully capture nuanced aspects of argumentation and discourse coherence \cite{dikli2006overview,madnani2018automated}. As machine learning techniques advanced, ensemble models and richer data-driven feature representations improved predictive robustness and generalization across prompts. Nevertheless, challenges related to construct validity, domain transferability, and interpretability persisted, highlighting the limitations of purely surface-level linguistic features \cite{williamson2012framework,lottridge2023psychometric,ramesh2022automated}.

Recent advances in deep learning and neural language models have fundamentally transformed automated scoring methodologies. Neural architectures allow systems to learn high-dimensional representations of text directly from raw essays, reducing dependence on manual feature engineering and enabling better modeling of semantic content and discourse-level structure. Recurrent neural networks and convolutional neural networks were among the earliest neural approaches shown to improve AES performance by capturing sequential and hierarchical language patterns \cite{taghipour2016neural,alikaniotis2016automatic,dong2017attention}. These models demonstrated that distributed representations of text could better capture relationships between sentences, paragraph structure, and thematic development. The development of transformer architectures further accelerated progress in text modeling by enabling more effective representation of long-range contextual dependencies and parallel processing of textual inputs \cite{vaswani2017attention}. Pretrained language models such as BERT demonstrated that contextual embeddings learned from large corpora could substantially enhance performance across a wide range of NLP tasks, including writing evaluation and semantic similarity estimation \cite{devlin2019bert}. These advances shifted AES research toward representation learning approaches that emphasize semantic understanding rather than handcrafted indicators. The emergence of large language models (LLMs) has extended these capabilities beyond representation learning into generative and evaluative functions. Foundation models trained on massive corpora exhibit strong zero-shot and few-shot learning abilities, enabling them to evaluate, summarize, and generate text in ways that increasingly resemble human language processing \cite{bommasani2021foundation}. Modern generative AI systems such as GPT-4 demonstrate advanced reasoning, summarization, and evaluative capacities that make them particularly promising for educational applications requiring nuanced text understanding \cite{openai2023gpt4,bubeck2023paper}. Unlike earlier AES pipelines that required task-specific model training and extensive labeled datasets, LLM-based systems can perform rubric-aligned scoring, provide explanatory feedback, and assist instructors in reviewing student work with minimal additional supervision. This shift reduces reliance on domain-specific feature engineering and enables the more flexible integration of assessment criteria into automated systems.

Generative AI has therefore introduced new possibilities for scalable writing assessment that extend beyond traditional scoring pipelines. LLMs can be integrated into AES frameworks to perform tasks such as adaptive summarization of long essays, semantic evaluation of argument quality, automated feedback generation, and rubric-guided explanation of scoring decisions. These capabilities are especially valuable in higher education settings, where rapid formative feedback can enhance learning outcomes, encourage revision practices, and reduce instructor workload \cite{naseer2024automated}. At the same time, the integration of generative AI raises important methodological and ethical considerations. From a psychometric perspective, AES systems must continue to satisfy standards of reliability, validity, fairness, and transparency before being adopted for educational decision-making \cite{williamson2012framework,lottridge2023psychometric}. Furthermore, as generative tools become widely accessible to students, distinguishing between authentic writing and AI-assisted composition becomes increasingly challenging. The availability of AI-assisted writing tools may blur the boundaries between student ability and machine support, raising new concerns about academic integrity and equitable assessment \cite{song2023enhancing,mao2024generative}. 

Recent research has begun exploring the direct use of LLMs for automated scoring tasks, demonstrating promising levels of agreement with human raters while also highlighting concerns about bias, interpretability, computational cost, and environmental impact \cite{mansour2024can,kasneci2023chatgpt}. Recent 2025--2026 studies further show that the field is moving in several directions at once: prompt-based and zero-shot LLM scoring, comparative scoring with pairwise preferences, bias analysis under holistic and analytic rubrics, graph-based multi-trait scoring, explainable scoring-and-feedback systems, and psychometrically grounded scoring-and-feedback frameworks \cite{cai2025rank,shibata2025lces,kucia2026llm,li2025gat,li2025ceaes,xia2026psyscore}. These developments emphasize the need for systematic investigation of how model design choices affect scoring performance, resource efficiency, and educational applicability. In particular, understanding trade-offs among model size, computational cost, and scoring accuracy is essential for institutions seeking practical deployment strategies. Within this evolving landscape, the present study examines how GPT-based generative models can be integrated into an automated essay-scoring framework using adaptive summarization and semantic embedding techniques. Specifically, we compare model variants to evaluate trade-offs among scoring accuracy, summarization quality, and computational efficiency, with the goal of informing practical model selection strategies for scalable, resource-efficient automated assessment systems.

The contribution of this study is therefore deliberately narrower than direct LLM-as-scorer research. Instead of asking whether an LLM can assign a final essay score directly, we examine whether generative summarization can act as an intermediate compression mechanism that makes long essays usable for fixed-length embedding and classifier pipelines while retaining interpretable handcrafted features from the original text. This framing distinguishes the proposed system from three adjacent approaches: naive truncation, which discards the tail of long essays; long-context encoders such as Longformer, which directly extend the input window but still require task-specific fine-tuning and computational resources; and direct zero-shot LLM scoring, which reduces training requirements but introduces prompt sensitivity and score-calibration challenges. The present study contributes an empirical analysis of the summarization-preprocessing path, the downstream scoring consequences of different GPT summarizers, and score-band differences in information preservation.

\section{Background}

Prior research in automated essay scoring has progressively shifted from rule-based and feature-driven approaches toward data-driven and neural methodologies. Early operational AES systems demonstrated that computational scoring could achieve levels of agreement comparable to those of human raters under controlled conditions. For example, Project Essay Grade (PEG) established that surface-level statistical indicators such as word frequency and essay length could predict writing quality with moderate success, achieving correlations with human raters comparable to inter-rater reliability \cite{page1966imminence,dikli2006overview}. Subsequent systems, such as e-rater, expanded this paradigm by incorporating grammar, usage, mechanics, and discourse-level features into regression models, allowing automated scores to reach human-machine agreement levels above $r=0.70$ in standardized testing environments \cite{attali2006automated,burstein2004automated}. These results confirmed that structured linguistic indicators could approximate holistic scoring criteria while supporting large-scale assessment operations.

The development of psycholinguistic feature extraction tools further improved AES performance. Coh-Metrix introduced theory-driven indices related to lexical diversity, cohesion, syntactic complexity, and readability, enabling deeper modeling of writing constructs \cite{graesser2004coh}. Studies applying Coh-Metrix features to large-scale language assessments reported substantial gains in scoring accuracy compared to surface-only models, particularly for evaluating discourse coherence and text organization \cite{latifi2021automated}. For instance, automated scoring systems integrating cohesion and syntactic indices demonstrated improved agreement with expert raters and better construct representation across grade levels. These findings highlighted the importance of aligning computational features with cognitive and linguistic theories of writing proficiency \cite{williamson2012framework,lottridge2023psychometric}.

Machine learning methods enhanced the predictive robustness of AES by enabling nonlinear modeling of complex linguistic patterns. Ensemble techniques such as random forests and gradient boosting improved generalization across prompts and reduced sensitivity to feature noise \cite{madnani2018automated}. Empirical evaluations showed that ensemble models outperformed single-model baselines and achieved performance competitive with operational AES platforms \cite{ramesh2022automated}. However, these approaches still relied heavily on handcrafted features and required extensive task-specific engineering, limiting adaptability across domains and writing contexts.

Neural network approaches marked a major methodological transition by enabling automatic representation learning from raw text. Recurrent neural networks and convolutional neural networks demonstrated improved ability to capture semantic relationships and discourse flow without explicit feature design \cite{alikaniotis2016automatic,taghipour2016neural}. Early neural AES systems achieved higher agreement with human raters than traditional feature-based models, with performance gains particularly evident for longer and more complex essays. Attention mechanisms further enhanced interpretability by identifying salient textual regions influencing scoring decisions \cite{dong2017attention}.

Transformer-based language models significantly advanced AES capabilities by improving contextual representation and long-range dependency modeling. Pretrained encoder architectures enabled transfer learning across writing prompts and assessment tasks, reducing the need for large labeled datasets \cite{devlin2019bert,liu2019roberta,he2021deberta}. Empirical studies have demonstrated that contextual embeddings improve both scoring accuracy and robustness to prompt variation, thereby supporting more scalable assessment pipelines.

Recent work has investigated the direct application of large language models to automated scoring tasks. Studies evaluating GPT-based scoring frameworks report promising human-machine agreement, with several experiments demonstrating near-human consistency under rubric-guided prompting \cite{mansour2024can,lee2024applying,pack2024large}. Chain-of-thought prompting has been shown to improve the transparency of evaluative reasoning and the stability of scoring by encouraging models to generate intermediate justification steps before assigning scores \cite{lee2024applying}. Fine-tuning strategies further enhance scoring reliability by adapting general-purpose models to domain-specific rubrics and assessment standards \cite{latif2024fine}.

However, emerging evidence also highlights limitations of LLM-based scoring. Variability across prompts, sensitivity to phrasing of instructions, and inconsistencies in repeated evaluations raise concerns about reliability for high-stakes use \cite{kasneci2023chatgpt,pack2024large}. Computational cost is another major constraint, as larger models require substantially more energy and infrastructure resources \cite{bommasani2021foundation}. Comparative analyses indicate that smaller, optimized variants often achieve performance close to that of large-scale models while significantly reducing computational overhead, suggesting important efficiency–performance trade-offs for practical deployment \cite{mansour2024can,lee2024applying}.

Overall, prior AES research demonstrates a trajectory toward increasingly sophisticated language understanding mechanisms, progressing from handcrafted feature engineering to neural representation learning and generative foundation models \cite{dikli2006overview,madnani2018automated,lottridge2023psychometric}. While modern systems achieve strong performance, challenges remain in balancing scoring accuracy, computational efficiency, interpretability, and psychometric validity. These considerations motivate a systematic investigation into how different model architectures and configurations influence both scoring outcomes and resource requirements.

\begin{table}[h]
\centering
\begin{tabular}{l c l}
\hline
\textbf{Model} & \textbf{Typical Max Input Length} & \textbf{Notes} \\
\hline
BERT      & 512 tokens & Standard encoder; common AES baseline \\
RoBERTa   & 512 tokens & Robustly optimized BERT variant \\
DeBERTa   & 512 tokens & Disentangled attention encoder \\
ELECTRA   & 512 tokens & Efficient pretraining via replaced token detection \\
\hline
\end{tabular}
\caption{Typical input-length limits of representative transformer models used in text scoring and long-document processing.}
\label{tab:token_limits}
\end{table}

Although modern encoder-based language models have substantially improved automated essay scoring, most standard pretrained architectures remain constrained by relatively short input windows. BERT was pretrained with sequence lengths up to 512 tokens \cite{devlin2019bert}, and RoBERTa likewise used a maximum sequence length of 512 tokens during pretraining \cite{liu2019roberta}. DeBERTa and ELECTRA are also commonly configured with a default maximum of 512 position embeddings \cite{he2021deberta,clark2020electra}. This limitation is especially important in AES, where long-form student essays frequently exceed the effective input length of these models. Because transformer self-attention scales quadratically with sequence length, processing long documents directly becomes computationally expensive and often infeasible with standard encoder architectures \cite{vaswani2017attention,beltagy2020longformer}. Long-document encoders such as Longformer mitigate this bottleneck by using sparse attention and can process substantially longer inputs \cite{beltagy2020longformer}. They are therefore an important baseline for future work. However, they do not eliminate the broader deployment question addressed here: many AES systems still depend on fixed-length embedding APIs or standard encoders, and institutions may prefer preprocessing strategies that can be attached to existing scoring pipelines without retraining a new long-context model. As a result, truncation may remove important discourse, evidence, and organizational cues relevant to essay quality evaluation, while summarization offers an alternative way to preserve compressed semantic content.

Generative AI provides a practical way to address this long-text bottleneck. Unlike encoder-only models that primarily transform input text into fixed representations, generative models can summarize long essays into shorter semantically salient versions while retaining core arguments, topical relevance, and discourse structure. This is particularly useful for AES because summarized texts can be compressed below encoder token limits without relying on naive truncation. In effect, generative summarization can serve as an information-preserving preprocessing step, enabling downstream embedding and classification models to process long essays more efficiently. This design is not presented as a replacement for Longformer-style modeling or direct LLM scoring. Rather, it is a complementary route for settings where the scoring architecture is already built around fixed-length embeddings, tabular features, and conventional supervised classifiers. In the present study, this rationale supports the use of GPT-based summarization as a bridge between long-form essays and encoder-based automated scoring models.

\begin{table}[h]
\centering
\begin{tabular}{l c c}
\hline
\textbf{Model} & \textbf{Input (per 1M)} & \textbf{Output (per 1M)} \\
\hline
GPT-5      & \$1.25 & \$10.00 \\
GPT-5 mini & \$0.25 & \$2.00 \\
GPT-5 nano & \$0.05 & \$0.40 \\
\hline
\end{tabular}
\caption{API pricing for GPT-5 model variants per 1M text tokens, based on OpenAI model documentation pages accessed on July 1, 2026.}
\label{tab:api_pricing}
\end{table}

As shown in Table~\ref{tab:api_pricing}, the API pricing structure of GPT model variants differs substantially across model sizes, creating important cost–performance trade-offs for large-scale applications. Larger-capacity models generally offer stronger language understanding and text generation capabilities, but they also incur significantly higher input and output token costs. In contrast, smaller variants offer more economical use with only modest performance reductions. For resource-intensive tasks such as training automated essay scoring systems--where large corpora must be processed iteratively for summarization, embedding generation, and model optimization--these pricing differences can translate into considerable variations in computational expenditure and processing time.

Therefore, model selection becomes not only a performance consideration but also a practical optimization problem. Choosing an appropriate model variant requires balancing scoring accuracy, computational efficiency, and budget constraints. In large-scale educational assessment scenarios, where efficiency, scalability, and sustainability are critical, selecting a cost-effective model can substantially reduce financial burden and energy consumption while still maintaining reliable automated scoring performance. This cost–efficiency perspective further motivates the comparative evaluation of different GPT model variants conducted in the present study.

This project is guided by the central research question of how generative AI-based summarization can improve the cost efficiency and scalability of automated essay scoring systems without compromising scoring reliability. Specifically, the study investigates whether using different generative model variants to compress long-form essays into semantically preserved summaries can reduce computational burden while maintaining effective representation of writing quality. By transforming lengthy essays into concise summaries, generative models may help mitigate token-length constraints of downstream embedding and classification architectures, thereby improving processing efficiency for large-scale assessment pipelines. To comprehensively evaluate this approach, the study adopts a hybrid feature perspective that integrates both summarized and non-summarized information sources. On one hand, summarized texts generated by generative AI models serve as inputs for semantic embedding, enabling efficient representation learning from condensed discourse. On the other hand, handcrafted linguistic features extracted from the original, unsummarized essays capture structural, lexical, syntactic, and readability characteristics that may be partially lost during summarization. By combining compressed semantic representations with interpretable surface-level linguistic indicators, the proposed framework aims to balance efficiency, information preservation, and scoring robustness.

Because recent AES benchmarks include strong alternatives such as NPCR, T-MES, CEAES, GAT-AES, and direct or comparative LLM scoring \cite{xie2022npcr,wang2025tmes,li2025ceaes,li2025gat,shibata2025lces}, the present experiments should not be read as establishing a new state of the art. Their purpose is to isolate a more specific question: under an otherwise fixed hybrid AES pipeline, how do different GPT summarizers affect summary fidelity, downstream agreement with human scores, and practical cost trade-offs? This controlled scope is important because a full SOTA benchmark would require retraining multiple architectures on identical splits and resources, which is beyond the current experimental evidence reported here.

Through this dual-representation strategy, the project seeks to assess how model size, summarization quality, and computational cost interact within automated scoring pipelines. The broader perspective of this research is that generative AI should not merely replace traditional scoring components but rather function as an adaptive preprocessing mechanism that enhances scalability while retaining critical linguistic signals. This integrated approach provides insights into practical model selection and system design for cost-effective, reliable, and large-scale educational assessment.

\section{Methodology}

\subsection{Dataset}
This study uses the dataset released for the Learning Agency Lab -- Automated Essay Scoring 2.0 competition hosted on Kaggle. The dataset was designed to advance research in automated essay scoring (AES) by providing a large-scale collection of student-written argumentative essays paired with expert human ratings. The public competition release contains approximately 24,000 essays composed in response to standardized writing prompts. In the experiments reported here, the available labeled training subset after preprocessing contains the score-band distribution shown in Table~\ref{tab:trainingset}. Each essay is evaluated using a holistic scoring rubric and assigned a single integer score ranging from 1 to 6, where higher scores indicate stronger overall writing quality. The scoring process reflects authentic educational assessment practices, making the dataset particularly suitable for modeling real-world AES systems \cite{kaggle_aes2}.

\begin{table}[h]
\centering
\begin{tabular}{c c c c c c c}
\hline
\textbf{Score Band} & \textbf{Count} & \textbf{Mean} & \textbf{Std} & \textbf{Min} & \textbf{Median} & \textbf{Max} \\
\hline
1 & 1002 & 267.4 & 107.8 & 150 & 233.5 & 902 \\
2 & 3779 & 259.5 & 95.7  & 150 & 235.0 & 1656 \\
3 & 5024 & 353.8 & 100.4 & 152 & 338.0 & 1213 \\
4 & 3140 & 476.1 & 110.1 & 228 & 458.0 & 1268 \\
5 & 776  & 627.4 & 129.9 & 364 & 610.5 & 1312 \\
6 & 124  & 766.0 & 160.7 & 497 & 740.0 & 1367 \\
\hline
\end{tabular}
\caption{Training dataset descriptive statistics by score band.}
\label{tab:trainingset}
\end{table}

The training dataset in Table~\ref{tab:trainingset} exhibits clear length variation across score bands, with higher-scoring essays generally substantially longer. As the score increases from Band 1 to Band 6, both the mean and median token counts rise steadily, indicating that stronger essays tend to provide more developed and detailed responses. Notably, the maximum lengths in the upper bands are extremely large (e.g., exceeding 1,300 tokens), meaning a considerable portion of high-scoring essays surpass the common 512-token limit used in many language model embedding pipelines. This has important implications for preprocessing, as truncation may disproportionately affect top-performing responses. At the same time, essay length is not strictly determined by score level: even lower-scoring bands (e.g., Bands 1 and 2) contain unusually long submissions, with maximum lengths reaching 902 and 1656 tokens, respectively. These results indicate substantial overlap in length distributions across score levels, suggesting that token management strategies must accommodate both high-performing long essays and unexpectedly lengthy lower-scoring responses.

From a machine learning perspective, the task is formulated as a supervised text scoring problem. Given the essay text as input, models must predict a score that aligns with human judgment. The dataset supports both regression-based approaches, where models predict continuous scores later rounded to integers, and ordinal classification approaches, which treat scores as ordered categories. Model performance in the competition is evaluated using Quadratic Weighted Kappa (QWK), a metric that measures agreement between predicted and human-assigned scores while penalizing larger disagreements more heavily. This evaluation design encourages models to approximate expert scoring behavior rather than simply minimizing numerical error.

Because the data are publicly available and de-identified, the present work did not involve recruiting participants, interacting with students, or collecting new personally identifiable information. The analysis was therefore treated as secondary analysis of public data and did not require additional ethics approval under the author's institutional practice for non-interventional public-data research. The author remains responsible for ensuring that this statement is consistent with the final journal and institutional requirements before publication.

\subsection{Word embedding}

\begin{figure}
    \centering
    \includegraphics[width=0.5\linewidth]{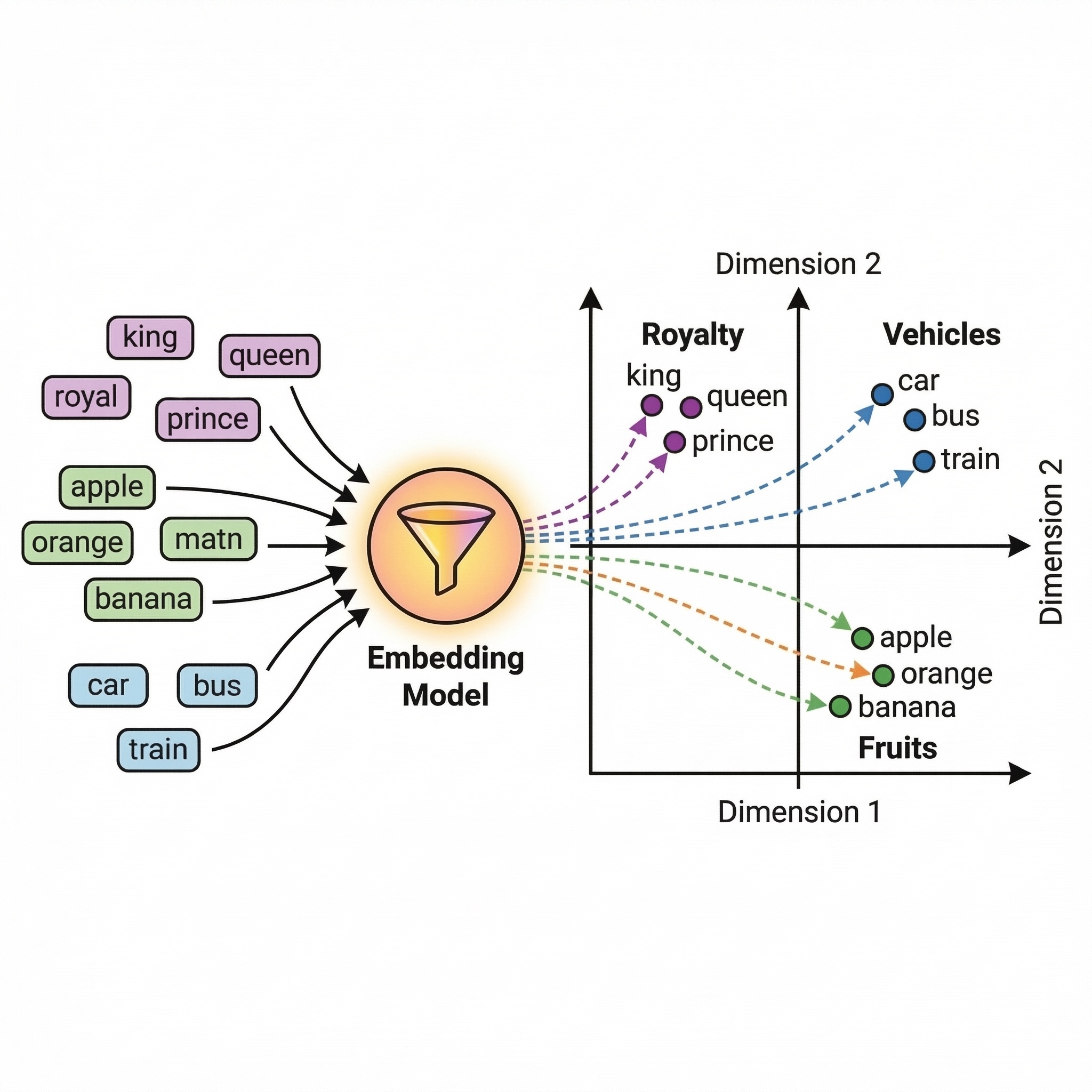}
    \caption{Conceptual illustration of word embedding, in which textual units are mapped into a continuous vector space so that semantically related expressions are located near one another.}
    \label{fig:1}
\end{figure}

As illustrated in Fig.~\ref{fig:1}, in natural language processing (NLP), word embeddings are dense vector representations that map discrete textual units (words, subwords, or sentences) into continuous numerical space. Unlike traditional one-hot encodings, which represent words as sparse vectors without semantic meaning, embeddings capture semantic similarity and contextual relationships between words by placing related terms closer together in vector space. For example, semantically related words such as \textit{education}, \textit{learning}, and \textit{teaching} are represented by vectors with small cosine distances. Modern embedding methods are typically learned through neural language models trained on large corpora. These models capture distributional semantics based on the principle that words appearing in similar contexts tend to share similar meanings. Embeddings serve as foundational inputs for downstream NLP tasks, including text classification, semantic similarity, clustering, and automated essay scoring (AES). By transforming raw text into semantically meaningful vectors, embedding models enable machine learning systems to process language quantitatively \cite{chowdhary2020natural}.

To obtain high-quality semantic representations of student essays, we employ the Alibaba Cloud Qwen3-Embedding-4B model, an instruction-tuned multilingual text embedding model released on Hugging Face. This model is built on the Qwen3 large language model architecture and is optimized to generate dense semantic embeddings for retrieval, similarity estimation, and clustering. Qwen3-Embedding-4B maps input text into a high-dimensional continuous vector space where semantically similar texts are positioned closer together. Compared to earlier static embedding approaches such as Word2Vec or GloVe, Qwen embeddings are contextualized, meaning that vector representations depend on the full sentence or document context. This property is particularly beneficial for automated essay scoring, where meaning emerges from discourse structure, argument coherence, and contextual language usage rather than isolated word choice. The model contains approximately 4 billion parameters, enabling it to capture nuanced semantic and syntactic patterns across long-form text. It supports multilingual inputs and is trained using large-scale contrastive learning objectives that align semantically similar texts while separating unrelated content in the embedding space. Such training improves the model’s ability to capture essay-level coherence and argumentative structure \cite{zhang2025qwen3}.

In this study, each essay is encoded into a fixed-length embedding vector using Qwen3-Embedding-4B. These vectors serve as high-level semantic features for downstream scoring models, allowing the system to evaluate latent qualities such as argument relevance, topic consistency, and semantic richness beyond surface-level lexical features.

\subsection{Automated scoring system}

\subsubsection{Text summarization}
In the previous study, we observed a length-control challenge in the GPT-based summarization modules (GPT-5, GPT-5 mini, and GPT-5 nano). Despite explicit token constraints, the generated summaries occasionally exceeded the maximum input limit of 512 tokens required by the downstream scoring model \cite{hua2025exploration}. To address this issue, we adopted an adaptive iterative summarization strategy. When a generated summary exceeded the allowable length, the model was prompted again with a stricter length specification set to 80\% of the previous target, progressively reducing the allowable output size until the summary met the token constraint. This controlled resummarization procedure ensured compliance with model input limits while preserving essential semantic content.

All summarization prompts were designed to enforce high semantic fidelity, requiring the model to retain the original essay’s arguments, intent, and domain-specific terminology while reducing redundancy through selective compression rather than aggressive paraphrasing. Compared to naive truncation approaches that risk discarding critical discourse elements, transformer-based summarization preserves logical structure and inter-sentence coherence by leveraging contextual language understanding. The adaptive length-control mechanism, therefore, improves both computational efficiency and representational consistency across essays of varying lengths, enabling fairer comparisons of automated scoring. This strategy builds upon prior findings demonstrating that controlled LLM-based summarization can enhance the robustness and comparability of automated essay scoring systems.

The summarization module was held constant across experimental conditions except for the GPT model variant. Each model received the same essay text, the same length-control instruction, and the same semantic-preservation instruction. The target output length was capped at 512 tokens to match the downstream embedding constraint. If a generated summary exceeded that cap, the iterative 80\% resummarization rule was applied until the output satisfied the length requirement. This design makes the comparison among GPT-5, GPT-5 mini, and GPT-5 nano attributable primarily to model-variant behavior rather than prompt or downstream-pipeline differences.

\subsubsection{Handcrafted features}
In addition to semantic embeddings derived from summarized essays, this study incorporates a set of handcrafted linguistic features extracted from the original, unsummarized texts to examine their complementary contribution to automated scoring performance. A total of 22 manually engineered features were designed across five dimensions to capture diverse aspects of writing quality. 
The first dimension consists of surface statistical features (7 features), including character count, word count, sentence count, paragraph count, average word length, average sentence length, and average paragraph length, which reflect fundamental text length and structural properties.
The second dimension includes lexical richness features (4 features), such as Type–Token Ratio (TTR), Measure of Textual Lexical Diversity (MTLD), proportion of advanced vocabulary, and hapax legomena ratio, which assess vocabulary diversity and sophistication. The third dimension comprises coherence and discourse features (6 features), including the total number of discourse connectives; counts of causal, contrastive, and progressive connectives; and the mean and standard deviation of semantic similarity between adjacent sentences, capturing logical organization and discourse flow. The fourth dimension contains readability features (3 features), including Flesch Reading Ease, Flesch–Kincaid Grade Level, and Gunning Fog Index, which quantify textual difficulty and cognitive load.
The final dimension includes error-related features (2 features), namely spelling error rate and number of grammatical errors, which reflect writing accuracy and adherence to language conventions.

By combining handcrafted features derived from the original essays with high-level semantic representations from summarized texts, the study evaluates whether traditional linguistic indicators provide complementary signals beyond neural embeddings. This hybrid feature framework enables a more comprehensive analysis of model performance and facilitates investigation into how surface-level writing characteristics interact with compressed semantic representations in automated essay scoring.

The handcrafted feature extraction pipeline was standardized across all summarization conditions. Surface statistics were computed from tokenized essay text; lexical richness features were computed from the ratio and distribution of unique word types; discourse-connective features were computed from curated connective categories; adjacent-sentence semantic similarity was computed using sentence-level embeddings; readability indices were computed using standard readability formulas; and error-related features were estimated using spelling and grammar checking tools. All 22 features were extracted from the original essay rather than the generated summary so that the hybrid model retained information that might be compressed or omitted during summarization.

\subsubsection{Classification system}

To model the relationship between linguistic features and essay quality, we employed two gradient-boosted decision-tree frameworks: XGBoost and LightGBM. These ensemble learning algorithms are well-suited for structured tabular features and have demonstrated strong performance in educational data mining and automated scoring tasks. XGBoost uses a regularized gradient-boosting framework that incorporates second-order optimization, shrinkage, and column subsampling to improve generalization and prevent overfitting. Its efficient parallelization and handling of sparse features make it particularly effective for high-dimensional handcrafted feature spaces \cite{chen2016xgboost}. LightGBM further improves computational efficiency through histogram-based decision tree learning and a leaf-wise tree growth strategy, which reduces memory usage and accelerates training while maintaining high predictive accuracy \cite{ke2017lightgbm}. Compared with traditional machine learning classifiers, both methods offer superior scalability and robustness for modeling nonlinear relationships between linguistic indicators and essay scores. By leveraging these complementary boosting frameworks, our study ensures reliable performance estimation and enables comparative evaluation of feature contributions across different model architectures.

All reported GPT-variant conditions used the same labeled data, handcrafted feature definitions, embedding model, and classifier family. The intended reproducible setup is as follows: essays are first split into training, validation, and test partitions using score-stratified sampling; summaries are generated only from essay text without access to human scores; embeddings and handcrafted features are concatenated after feature scaling; classifier hyperparameters are selected on the validation partition; and the final QWK is computed on the held-out test partition. For XGBoost and LightGBM, the relevant tuned hyperparameters include number of estimators, learning rate, maximum depth or number of leaves, subsampling ratio, column-sampling ratio, and regularization strength. This shared setup is important because the purpose of the comparison is to evaluate the effect of the summarization model, not differences in training data, feature sets, or classifier resources.

\begin{figure}
    \centering
    \includegraphics[width=0.9\linewidth]{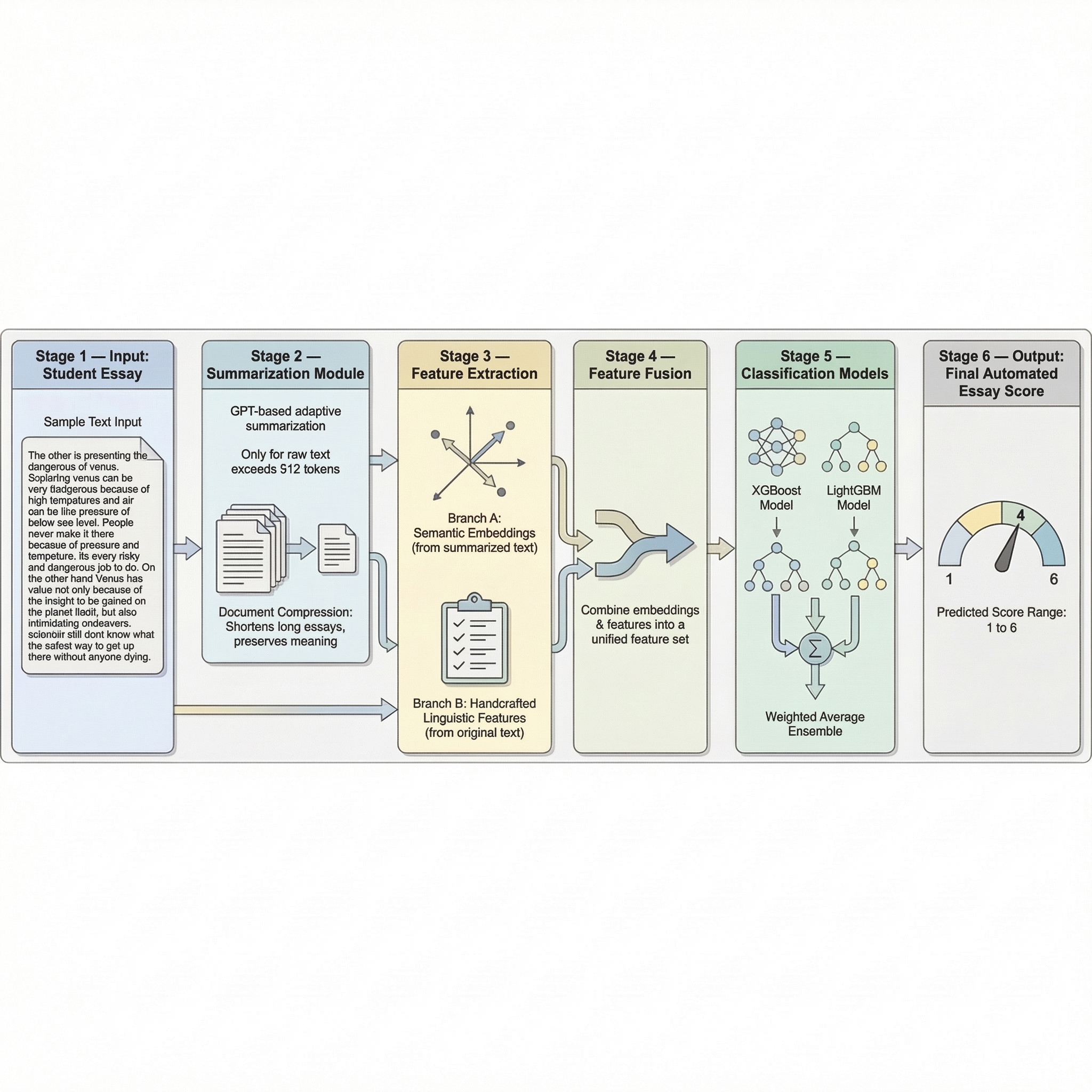}
    \caption{Proposed AES pipeline. Original essays are adaptively summarized by GPT variants, encoded with Qwen3-Embedding-4B, combined with 22 handcrafted linguistic features extracted from the original essays, and scored using gradient-boosted classifiers.}
    \label{fig:2}
\end{figure}

As illustrated in Fig.~\ref{fig:2}, the proposed automated essay scoring system integrates neural text processing with traditional linguistic feature engineering in a unified framework. Raw student essays are first processed through an adaptive GPT-based summarization module to ensure length compliance while preserving semantic integrity. The summarized texts are then encoded into dense semantic embeddings, while the original essays are simultaneously analyzed to extract handcrafted linguistic features that capture structural, lexical, discourse, readability, and accuracy characteristics. These complementary feature sets are subsequently fused into a unified representation and fed into gradient boosting classifiers (XGBoost and LightGBM) to model nonlinear relationships between writing attributes and essay quality. This hybrid pipeline enables efficient handling of long-form texts while leveraging both high-level semantic representations and interpretable linguistic indicators, resulting in a robust and scalable automated scoring framework.

\subsection{Baseline and ablation design}

The reported results focus on within-pipeline GPT summarizer comparison. A complete AES benchmark should additionally include five baseline families: (1) raw-text truncation to 512 tokens, (2) long-document transformer baselines such as Longformer, (3) summary-embedding-only models, (4) handcrafted-feature-only models, and (5) direct zero-shot or few-shot LLM scoring with models such as GPT, Claude, Gemini, Llama, and Qwen. These baselines are necessary because they test distinct claims. Truncation tests whether summarization preserves information that would otherwise be discarded. Longformer tests whether compression is preferable to direct long-context modeling. Embedding-only and handcrafted-only ablations quantify the independent contribution of each feature source. Direct LLM scoring tests whether a generative model should score essays directly rather than serve as a preprocessing component.

Because the present study has not executed all of these baselines, they are not reported as completed experiments. Instead, they define the ablation protocol required for a stronger benchmark. The core ablation matrix should compare the full hybrid model against: no summarization with truncation, GPT summary embedding without handcrafted features, handcrafted features without summary embedding, alternative classifier families, and direct LLM scoring. The same data splits, scoring rubric, QWK calculation, and computational budget accounting should be applied to every condition. This design would directly address whether the observed QWK values are attributable to the summarization module, the embedding model, the handcrafted features, the classifier, or their interaction.

\subsection{Automated scoring system performance evaluation}

 The QWK score is defined as follows: 
    \begin{equation}
        \kappa = 1 - \frac{\sum_{i,j} w_{i,j} O_{i,j}}{\sum_{i,j} w_{i,j} E_{i,j}},\label{eq:05}
     \end{equation}
    where \(i\) and \(j\) index the human and predicted score categories, \(w_{i,j}\) denotes the quadratic penalty assigned to disagreement between categories \(i\) and \(j\), \(O_{i,j}\) is the observed confusion-matrix frequency, and \(E_{i,j}\) is the expected frequency under chance agreement.

Quadratic Weighted Kappa (QWK) is used to evaluate the agreement between the automated scoring system and human raters for ordinal essay scores ranging from 1 to 6. Unlike standard accuracy or regression-based metrics, QWK accounts for the ordered nature of score categories and penalizes larger discrepancies more heavily than smaller ones. For example, predicting a score of 5 for an essay rated 6 by humans is considered a minor disagreement, whereas predicting 1 instead of 6 reflects a severe mismatch. QWK measures the extent to which the observed agreement between predicted and human scores exceeds the agreement expected by chance, while applying quadratic weights to emphasize the magnitude of rating differences. This design makes the metric particularly suitable for educational assessment tasks, where preserving relative score ordering and approximating human judgment are more important than minimizing raw numerical error. A QWK value of 1 indicates perfect agreement, values near 0 indicate chance-level agreement, and negative values indicate systematic disagreement \cite{doewes2023evaluating}.

\subsection{Summarized text quality evaluation}

\subsubsection{Basic similarity metrics}

ROUGE-N measures n-gram overlap between a reference text and a generated summary:

\begin{equation}
\text{ROUGE-N} =
\frac{\sum_{g \in \text{Ref}} \min \left( \text{Count}_{\text{cand}}(g), \text{Count}_{\text{ref}}(g) \right)}
{\sum_{g \in \text{Ref}} \text{Count}_{\text{ref}}(g)}
\end{equation}

The F1-score formulation is:

\begin{equation}
F_1 =
\frac{2 \cdot \text{Precision} \cdot \text{Recall}}
{\text{Precision} + \text{Recall}}
\end{equation}

ROUGE metrics evaluate lexical overlap between the original essays and their summaries. ROUGE-1 measures unigram (word-level) overlap, ROUGE-2 captures bigram (phrase-level) co-occurrence, and ROUGE-L evaluates the longest common subsequence to reflect structural similarity. The F1 formulation balances precision and recall, providing a comprehensive estimate of how much surface-level content from the source text is preserved. These metrics are widely used in summarization evaluation to assess lexical-level content retention \cite{lin2004rouge}.
In the ROUGE-N equation, \(g\) denotes an \(n\)-gram, \(\text{Ref}\) denotes the reference or source-side comparison set, \(\text{cand}\) denotes the candidate summary, and \(\text{Count}(\cdot)\) denotes the number of occurrences of an \(n\)-gram. Precision is computed as the proportion of candidate-summary \(n\)-grams that overlap with the reference, while recall is the proportion of reference \(n\)-grams retained in the candidate summary.

\begin{equation}
\text{CosineSim}(\mathbf{A}, \mathbf{B}) =
\frac{\mathbf{A} \cdot \mathbf{B}}
{\|\mathbf{A}\| \|\mathbf{B}\|}
\end{equation}

As shown above, cosine similarity measures semantic similarity between the original text and its summary by computing the cosine of the angle between their vector embeddings in high-dimensional space. Here, \(\mathbf{A}\) and \(\mathbf{B}\) represent the embedding vectors of the source essay and generated summary, respectively, \(\mathbf{A}\cdot\mathbf{B}\) is their dot product, and \(\|\cdot\|\) denotes vector magnitude. Unlike lexical-overlap metrics, cosine similarity captures meaning preservation even when different wording or paraphrasing is used. A higher cosine value indicates stronger semantic alignment, making this metric particularly suitable for evaluating abstractive summarization, where surface forms may differ while core meanings remain consistent \cite{gunawan2018implementation}.

\subsubsection{Information preservation metrics}

\begin{equation}
\text{Compression Ratio} =
\frac{\text{Length}_{\text{summary}}}
{\text{Length}_{\text{source}}}
\end{equation}

Compression ratio quantifies the extent to which the original essay is condensed during summarization, typically computed as the ratio between summary length and source text length. Length is measured using the same tokenization procedure for both source and summary. This metric reflects summarization efficiency by indicating how much content is reduced while maintaining essential information. An appropriate compression ratio ensures summaries are concise without excessive information loss \cite{jing2000sentence}.

\begin{equation}
\text{Entity Coverage} =
\frac{|\text{Entities}_{\text{summary}} \cap \text{Entities}_{\text{source}}|}
{|\text{Entities}_{\text{source}}|}
\end{equation}

Entity coverage measures the proportion of named entities (e.g., persons, organizations, locations, dates) retained in the summary relative to the original text. \(\text{Entities}_{\text{source}}\) and \(\text{Entities}_{\text{summary}}\) denote the sets of named entities extracted from the source essay and generated summary. Since named entities often represent key factual components, this metric assesses whether important information is preserved during compression. Higher entity coverage indicates better retention of factual details \cite{sang2003introduction}.

\begin{equation}
\text{KeyPhraseOverlap} =
\frac{|\text{KeyPhrases}_{\text{summary}} \cap \text{KeyPhrases}_{\text{source}}|}
{|\text{KeyPhrases}_{\text{source}}|}
\end{equation}

Key phrase overlap evaluates the extent to which important topic-representative phrases remain in the summary. \(\text{KeyPhrases}_{\text{source}}\) and \(\text{KeyPhrases}_{\text{summary}}\) denote the extracted keyphrase sets from the original essay and summary, respectively. These phrases are typically extracted using statistical or linguistic methods and represent central ideas of the text. Preserving key phrases ensures that summaries maintain the primary themes and argumentative focus of the original essays \cite{mihalcea2004textrank}.

\subsubsection{Quality control metrics}

\begin{equation}
\text{FactScore} =
\frac{\text{\# Consistent Sentences}}
{\text{Total Sentences}}
\end{equation}

\begin{equation}
\text{QuestEval} =
\frac{1}{Q} \sum_{i=1}^{Q}
\text{Sim}(A_i^{\text{source}}, A_i^{\text{summary}})
\end{equation}

FactScore and QuestEval are metrics designed to assess factual consistency between the summary and the source text. In the FactScore equation, consistent sentences are summary sentences judged to be supported by the source essay. In the QuestEval equation, \(Q\) denotes the number of generated question-answer pairs, \(A_i^{\text{source}}\) denotes the answer inferred from the source essay for question \(i\), \(A_i^{\text{summary}}\) denotes the answer inferred from the generated summary, and \(\text{Sim}(\cdot)\) denotes semantic similarity between the two answers. These metrics evaluate whether generated statements are logically supported by the original content, helping detect hallucinated or distorted information introduced during summarization \cite{kryscinski2020evaluating,scialom2021questeval}.

\begin{equation}
\text{Redundancy} =
\frac{1}{N(N-1)} \sum_{i \ne j}
\text{Sim}(s_i, s_j)
\end{equation}

Redundancy score measures the degree of repetitive information within a summary by identifying duplicated phrases or semantically overlapping sentences. Here, \(N\) denotes the number of sentences in a generated summary, \(s_i\) and \(s_j\) denote two distinct summary sentences, and \(\text{Sim}(s_i,s_j)\) denotes their semantic similarity. High redundancy reduces informational efficiency and affects readability. This metric ensures that summaries remain concise, informative, and free of unnecessary repetition \cite{see2017get}.

\section{Results}

\begin{table}[h]
\centering
\begin{tabular}{l c}
\hline
\textbf{Model Name} & \textbf{QWK Score} \\
\hline
GPT-5        & 0.8350 \\
GPT-5 mini   & \textcolor{red}{0.8435} \\
GPT-5 nano   & 0.8332 \\
\hline
\end{tabular}
\caption{Downstream AES performance after using each GPT-5 variant as the summarization module. QWK evaluates agreement between automated scores and human ratings; it is not used as a summarization-quality metric.}
\label{tab:model_performance}
\end{table}

The comparative performance of the evaluated model variants is summarized in Table~\ref{tab:model_performance}. The table reports downstream AES performance, not summarization quality. In other words, QWK is used here to test whether summaries generated by each GPT variant remain useful for the final scoring model. Among the three configurations, GPT-5 mini achieved the highest agreement with human ratings, obtaining a QWK score of 0.8435. The full GPT-5 model followed closely with a QWK of 0.8350, while GPT-5 nano produced a slightly lower score of 0.8332. Although performance differences are relatively small, the results suggest that the mini variant provides the most favorable balance between scoring accuracy and model efficiency. This finding indicates that smaller, optimized model variants can achieve performance comparable to larger models while potentially offering advantages in computational cost and deployment flexibility.

\begin{table}[h]
\centering
\begin{tabular}{l c c c}
\hline
\textbf{Metric} & \textbf{GPT-5} & \textbf{GPT-5 mini} & \textbf{GPT-5 nano} \\
\hline
ROUGE-1 F1 Score & \textcolor{red}{0.9261 $\pm$ 0.1370} & 0.9137 $\pm$ 0.1530 & 0.9001 $\pm$ 0.1729 \\
ROUGE-2 F1 Score & \textcolor{red}{0.8971 $\pm$ 0.1869} & 0.8657 $\pm$ 0.2331 & 0.8340 $\pm$ 0.2792 \\
ROUGE-L F1 Score & \textcolor{red}{0.9185 $\pm$ 0.1507} & 0.8978 $\pm$ 0.1812 & 0.8757 $\pm$ 0.2151 \\
Semantic Similarity & \textcolor{red}{0.9843 $\pm$ 0.0423} & 0.9800 $\pm$ 0.0438 & 0.9639 $\pm$ 0.0748 \\
Compression Ratio & 0.8933 $\pm$ 0.1907 & 0.8940 $\pm$ 0.1852 & \textcolor{red}{0.9023 $\pm$ 0.1732} \\
Entity Coverage & \textcolor{red}{0.9425 $\pm$ 0.1360} & 0.9330 $\pm$ 0.1493 & 0.8894 $\pm$ 0.2130 \\
Keyphrase Overlap & \textcolor{red}{0.8957 $\pm$ 0.2026} & 0.8923 $\pm$ 0.2028 & 0.8782 $\pm$ 0.2204 \\
Redundancy Score & \textcolor{red}{0.0333 $\pm$ 0.0357} & 0.0328 $\pm$ 0.0354 & 0.0329 $\pm$ 0.0354 \\
\hline
\end{tabular}
\caption{Summary statistics of evaluation metrics across GPT-5 model variants.}
\label{tab:summary_metrics}
\end{table}

The comparative evaluation results across automatic summarization quality metrics are presented in Table~\ref{tab:summary_metrics}. Overall, the full GPT-5 model demonstrates the strongest performance across most lexical-overlap and semantic-preservation measures. It achieves the highest mean scores for ROUGE-1, ROUGE-2, and ROUGE-L, indicating superior retention of surface-level lexical content relative to original essays. Similarly, GPT-5 attains the best performance in semantic similarity, entity coverage, and keyphrase overlap, suggesting that it more effectively preserves core meaning, salient entities, and important conceptual elements. These advantages reflect the stronger representational capacity of the larger model, which enables it to capture both fine-grained linguistic details and broader contextual relationships.

GPT-5 mini follows closely behind the full model, with only marginal decreases across nearly all evaluation dimensions. The relatively small performance gap indicates that the mini variant retains much of the linguistic and semantic modeling capabilities of GPT-5 while using fewer computational resources. This consistency suggests that architectural optimizations and parameter scaling allow smaller models to approximate the performance of larger counterparts on structured summarization tasks. In contrast, GPT-5 nano exhibits a more pronounced decline in lexical overlap and entity-based metrics, particularly in ROUGE-2 and entity coverage, implying weaker preservation of detailed phrasal information and important content units. Nevertheless, the nano variant achieves the highest compression ratio, indicating that it produces more concise summaries. This suggests a stronger tendency toward aggressive content condensation, which may improve brevity but also increase the risk of information loss. Despite these differences, redundancy scores remain low and nearly identical across all model variants, demonstrating comparable effectiveness in avoiding repetitive content.

Performance variability, as reflected by the reported standard deviations, remains moderate across models and metrics, indicating relatively stable summarization behavior. Notably, semantic similarity shows the lowest variability, suggesting that all model variants consistently maintain core meaning even when surface-level lexical overlap varies. Taken together, these findings highlight a clear performance–efficiency trade-off: while the full GPT-5 model provides the most comprehensive content retention and overall quality, smaller variants—particularly GPT-5 mini—offer highly competitive performance with improved computational efficiency, making them practical alternatives for large-scale automated summarization pipelines.

\begin{figure}[h]
    \centering
    \includegraphics[width=0.8\linewidth]{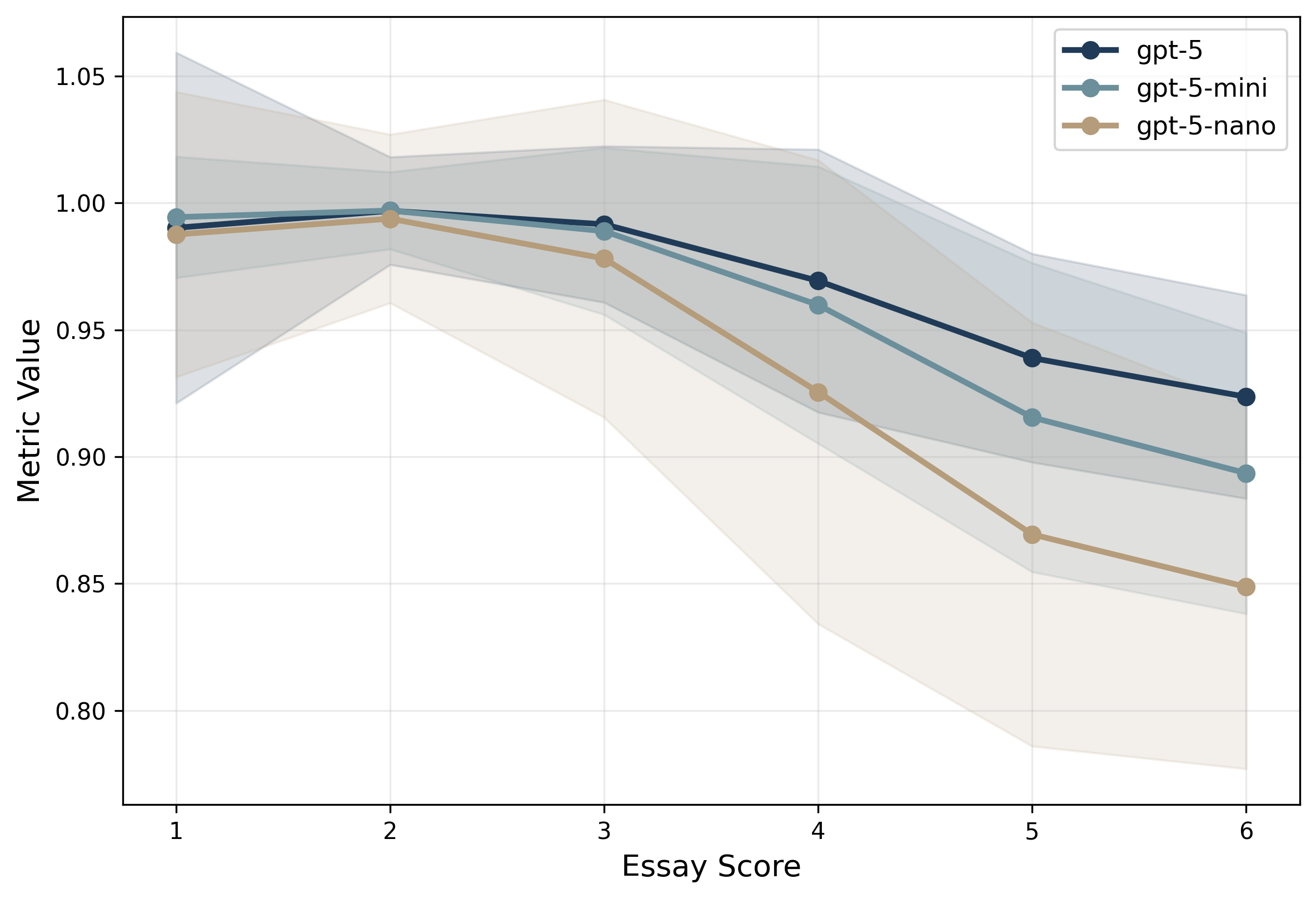}
\caption{Semantic similarity across essay score levels for different GPT model variants. Lines represent score-band means; the corresponding score-band sample sizes are reported in Table~\ref{tab:trainingset}.}
    \label{fig:semantic_similarity_score}
\end{figure}

Figure~\ref{fig:semantic_similarity_score} illustrates how semantic similarity varies across essay score levels for different GPT model variants. Overall, semantic similarity shows a gradual decline as essay scores increase, indicating that summaries of higher-scoring essays become progressively more abstractive and less lexically aligned with the source text. Among the models, GPT-5 consistently achieves the highest semantic similarity across all score levels, suggesting a stronger capability to preserve core meaning. GPT-5 mini follows closely, while GPT-5 nano shows a more pronounced decline at higher score bands. The widening performance gap at score levels 4–6 suggests that smaller models may struggle to maintain semantic fidelity when summarizing more complex or information-dense essays.

\begin{figure}[h]
    \centering
    \includegraphics[width=0.8\linewidth]{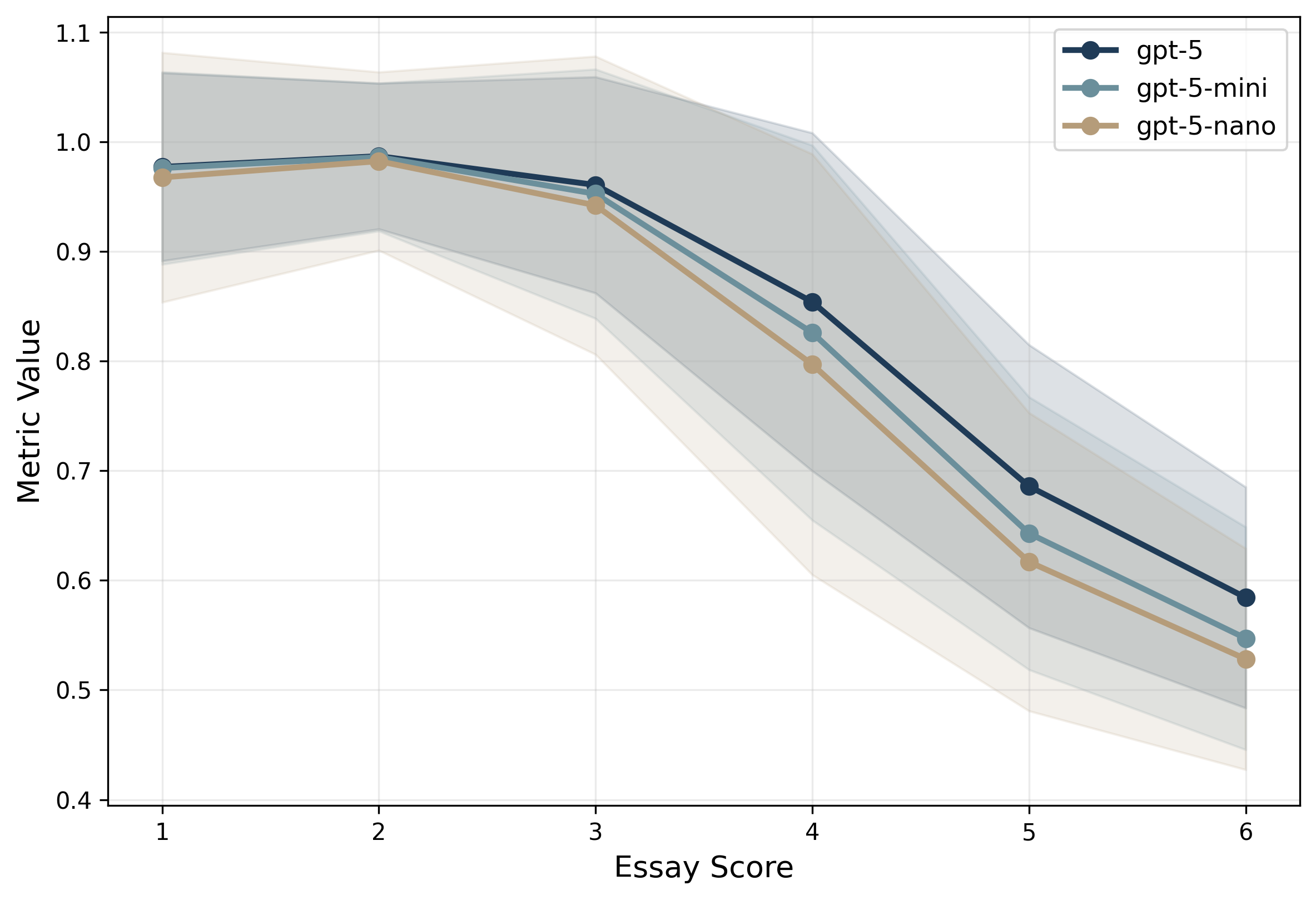}
\caption{ROUGE-1 F1 score across essay score levels for different GPT model variants. Lines represent score-band means; the corresponding score-band sample sizes are reported in Table~\ref{tab:trainingset}.}
    \label{fig:rouge1_score}
\end{figure}

Figure~\ref{fig:rouge1_score} presents the ROUGE-1 F1 score trends across essay score levels. A clear downward trajectory is observed as essay scores increase, indicating reduced lexical overlap between generated summaries and reference texts for higher-quality essays. This pattern suggests that high-scoring essays tend to require more abstractive summarization, which naturally lowers n-gram overlap. GPT-5 maintains the strongest lexical alignment throughout all score bands, while GPT-5 mini demonstrates comparable but slightly lower performance. GPT-5 nano shows the largest decrease, particularly beyond score level 4, implying that smaller-capacity models may have greater difficulty preserving surface-level textual features when handling more sophisticated content.

\begin{figure}[h]
    \centering
    \includegraphics[width=0.8\linewidth]{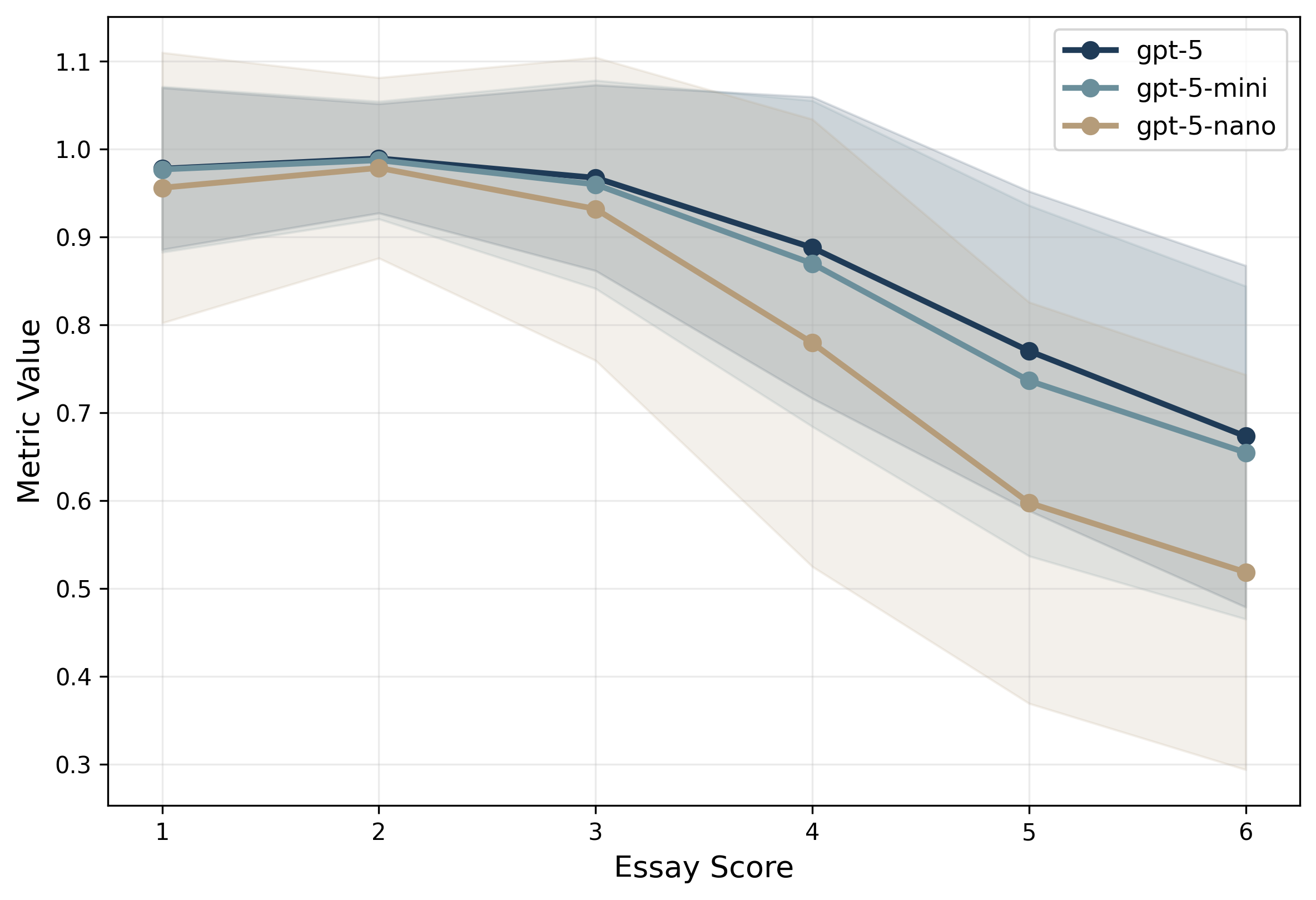}
\caption{Entity coverage across essay score levels for different GPT model variants. Lines represent score-band means; the corresponding score-band sample sizes are reported in Table~\ref{tab:trainingset}.}
    \label{fig:entity_coverage_score}
\end{figure}

Figure~\ref{fig:entity_coverage_score} shows entity coverage performance across essay score levels. Similar to other content-preservation metrics, entity coverage declines steadily as essay scores increase, indicating that summaries of more advanced essays tend to omit a greater proportion of named entities and key informational elements. GPT-5 consistently retains the highest proportion of entities, suggesting a stronger capability to preserve important factual content. GPT-5 mini performs competitively but shows a slightly steeper decline in higher score bands. GPT-5 nano shows the most substantial reduction, particularly at score levels 5 and 6, suggesting that smaller models may struggle to maintain detailed information when summarizing complex texts.

\begin{figure}[h]
    \centering
    \includegraphics[width=0.8\linewidth]{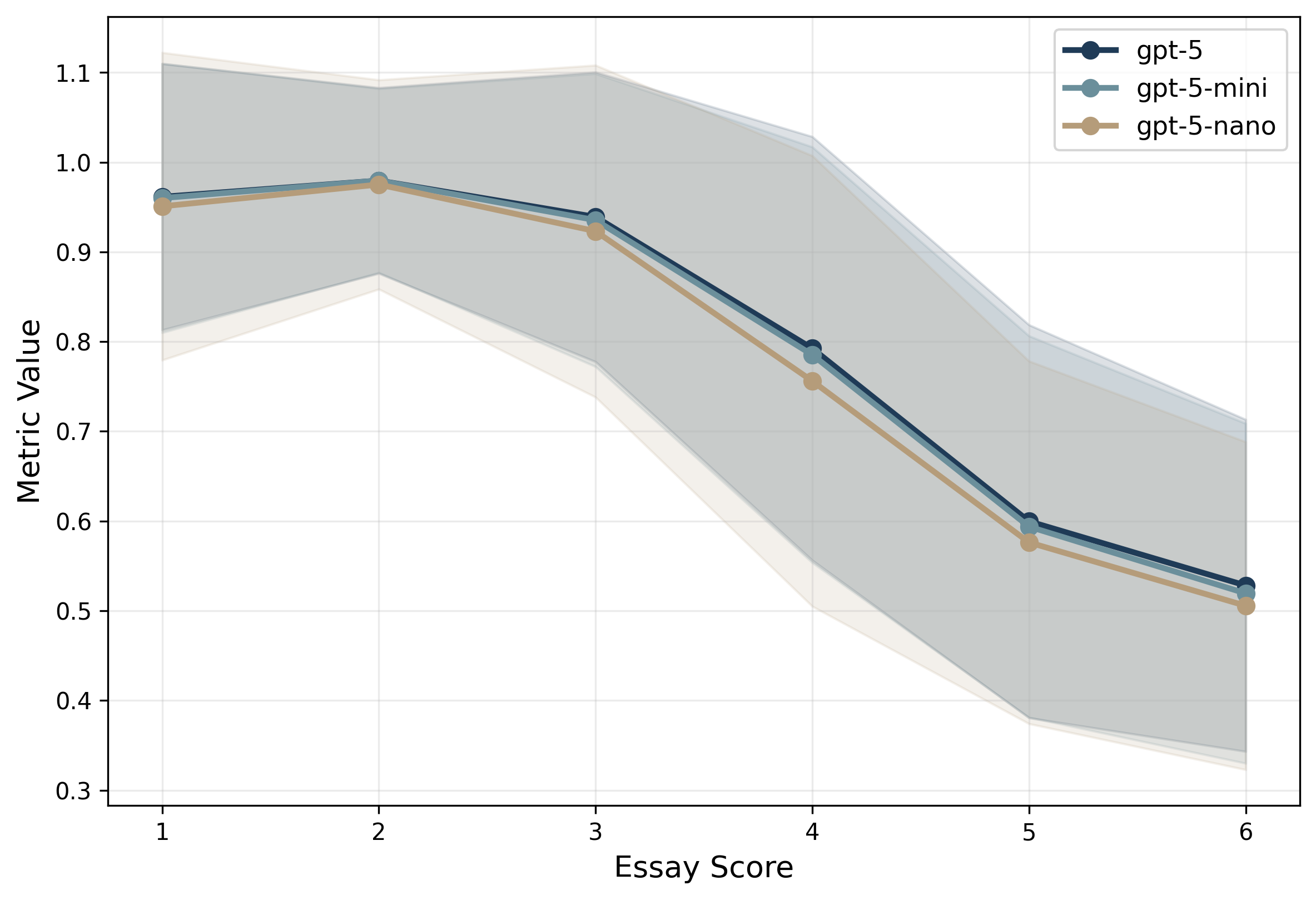}
\caption{Keyphrase overlap across essay score levels for different GPT model variants. Lines represent score-band means; the corresponding score-band sample sizes are reported in Table~\ref{tab:trainingset}.}
    \label{fig:keyphrase_overlap_score}
\end{figure}

Figure~\ref{fig:keyphrase_overlap_score} depicts the trend of keyphrase overlap across essay score levels. A consistent decreasing pattern is observed as essay scores increase, indicating that summaries of higher-scoring essays contain fewer shared salient phrases with the reference summaries. This suggests that advanced essays encourage more concept-level abstraction rather than direct phrase-level extraction. GPT-5 achieves the highest keyphrase retention across nearly all score levels, while GPT-5 mini shows similar performance with minor reductions. GPT-5 nano demonstrates a more pronounced drop at higher score bands, highlighting limitations in capturing central topical phrases when processing complex inputs. The relatively parallel trends across models suggest that keyphrase overlap is strongly influenced by essay complexity rather than model architecture alone.

\begin{figure}
    \centering
    \includegraphics[width=0.9\linewidth]{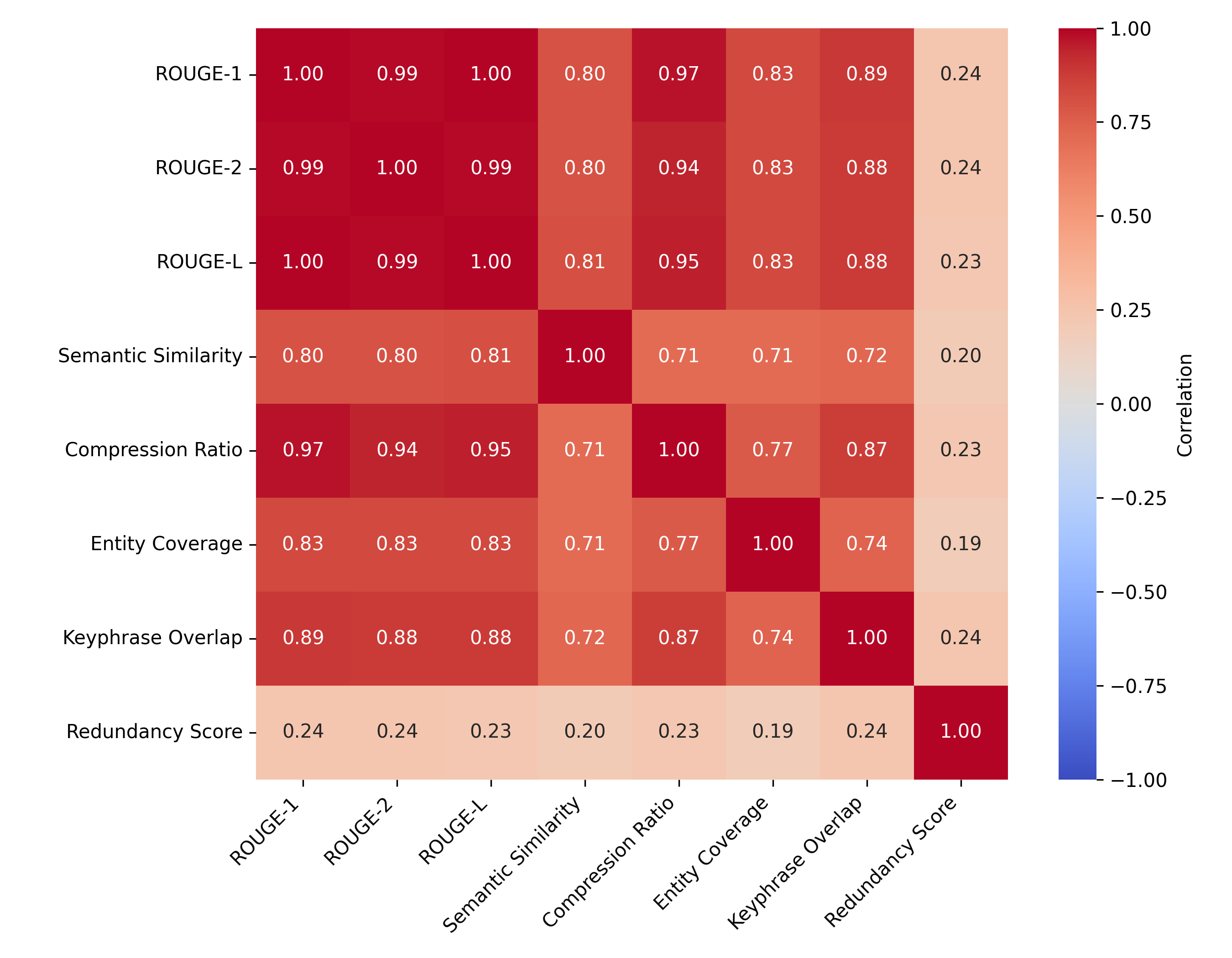}
\caption{Correlation matrix among summarization evaluation metrics, computed across the evaluated summary-level metric values.}
    \label{fig:metric_correlation}
\end{figure}

The correlation analysis among evaluation metrics is presented in Fig.~\ref{fig:metric_correlation}. Overall, the results show strong positive relationships among most content-preservation and overlap-based measures. In particular, ROUGE-1, ROUGE-2, and ROUGE-L are almost perfectly correlated with one another ($r=0.99$--$1.00$), indicating that these lexical overlap metrics capture highly similar aspects of summary quality. Compression ratio also shows strong positive correlations with the ROUGE metrics ($r=0.94$--$0.97$), suggesting that summaries with better lexical alignment to the reference also tend to retain a greater proportion of source information. Semantic similarity is moderately to strongly correlated with ROUGE and content coverage measures ($r=0.71$--$0.81$), indicating that semantic preservation is related to lexical overlap but still captures a somewhat distinct dimension of quality. Entity coverage and keyphrase overlap also show substantial positive correlations with other metrics, particularly ROUGE and the compression ratio, supporting their role as indicators of information retention. In contrast, the redundancy score has only weak positive correlations with all other measures ($r=0.19$--$0.24$), implying that repetitiveness is relatively independent from content preservation and semantic fidelity. Taken together, these findings suggest that most evaluation metrics converge on a shared notion of summary quality, whereas redundancy provides complementary information about stylistic efficiency rather than content alignment.

\section{Summary \& Discussion}

This study compared three GPT-based summarization variants—GPT-5, GPT-5 mini, and GPT-5 nano—within an automated essay scoring pipeline that combines summarized-text embeddings with handcrafted linguistic features. The overall scoring results show that GPT-5 mini achieved the highest agreement with human raters, with a QWK score of 0.8435, slightly outperforming GPT-5 and GPT-5 nano (Table~\ref{tab:model_performance}). This finding becomes especially meaningful when considered alongside the API pricing comparison (Table~\ref{tab:api_pricing}). The most expensive model did not yield the best downstream scoring performance. Although GPT-5 has the highest per-token cost, its scale advantage did not translate into a meaningful improvement in automated scoring accuracy. Instead, the relatively small performance differences across variants suggest that model cost and AES effectiveness are not directly proportional in this task setting.

The summarization quality results provide a more detailed picture of this trade-off. Across most text quality metrics, GPT-5 produced the strongest average performance, especially on ROUGE-1, ROUGE-2, ROUGE-L, semantic similarity, entity coverage, and keyphrase overlap (Table~\ref{tab:summary_metrics}). GPT-5 mini followed very closely across nearly all measures, while GPT-5 nano generally showed somewhat lower content-preservation performance but a higher compression ratio. These results suggest that the largest model preserved slightly more lexical and semantic information overall, but the magnitude of its advantage over GPT-5 mini was modest. In practical terms, GPT-5 mini appears to offer the best balance between summary quality, downstream AES performance, and computational efficiency, making it particularly suitable for large-scale automated scoring applications.

The four score-level figures reveal an additional and highly consistent pattern across models. Semantic similarity, ROUGE-1 F1, entity coverage, and keyphrase overlap all decline as essay score increases (Figs.~\ref{fig:semantic_similarity_score}--\ref{fig:keyphrase_overlap_score}). This indicates that summaries generated for higher-scoring essays tend to preserve less of the source text’s lexical content, semantic meaning, named entities, and central phrases than summaries generated for lower-scoring essays. GPT-5 remains the strongest model across most score levels, but all three variants exhibit downward trends as essay complexity increases. The decline becomes especially noticeable in the upper score bands, where GPT-5 nano drops the most sharply, and GPT-5 mini also shows greater separation from GPT-5. These patterns suggest that summarizing high-quality essays is systematically more challenging than summarizing lower-quality essays.

This trend is consistent with the training data statistics (Table~\ref{tab:trainingset}). Higher-scoring essays are substantially longer on average, and many exceed the 512-token threshold commonly used in embedding pipelines. As a result, the summarization module must compress far more information for strong essays than for weaker ones. This likely increases the risk of information loss during preprocessing, including omission of key entities, phrases, and semantic structure. By contrast, lower-scoring essays are typically shorter and require less aggressive compression. The observed drop in summarization quality for higher-score essays, therefore, reflects not only differences in model capacity but also structural challenges inherent in the dataset itself.

Finally, the correlation analysis clarifies how evaluation metrics relate to one another. ROUGE-1, ROUGE-2, and ROUGE-L are almost perfectly correlated, indicating that these overlap-based metrics capture nearly the same dimension of summary quality (Fig.~\ref{fig:metric_correlation}). Compression ratio, entity coverage, and keyphrase overlap also show strong positive relationships with the ROUGE metrics, while semantic similarity is moderately to strongly correlated with the other content-preservation measures. In contrast, redundancy exhibits only weak correlations with all other metrics, suggesting that repetitiveness represents a more independent stylistic dimension. Taken together, these results indicate that the evaluation framework is internally coherent and that most metrics converge on a shared notion of summary quality centered on lexical, semantic, and informational preservation under compression.

\section{Conclusion \& Future Work}

One of the most important findings of this study is that selecting the appropriate model variant can improve the practicality of automated essay scoring without requiring the most expensive language model. GPT-5 mini achieved the highest overall QWK score (Table~\ref{tab:model_performance}) while costing substantially less than the full GPT-5 model (Table~\ref{tab:api_pricing}). This demonstrates that superior AES performance does not necessarily result from choosing the largest or most resource-intensive model. Instead, performance appears to depend on how well a summarization model balances information preservation, compression behavior, and compatibility with the downstream scoring system. This insight has direct implications for real-world deployment. In large-scale educational settings, where thousands or millions of essays must be processed, a lower-cost model with near-equivalent or even superior performance can substantially reduce computational time, energy consumption, and financial burden.

A plausible explanation for the strong performance of GPT-5 mini is that it may generate summaries that are more efficiently aligned with the needs of the scoring model. Although GPT-5 achieved slightly better average results on most summarization-quality metrics (Table~\ref{tab:summary_metrics}), these incremental gains did not translate into the best final scoring performance. GPT-5 mini may preserve the most relevant content while avoiding unnecessary detail that a larger model might retain. In contrast, GPT-5 nano appears to compress more aggressively, as reflected by its higher compression ratio and lower entity and keyphrase retention (Table~\ref{tab:summary_metrics}), which may reduce the usefulness of the resulting summary representations for AES.

Another key finding is that summarization quality declines for higher-scoring essays across all three models (Figs.~\ref{fig:semantic_similarity_score}--\ref{fig:keyphrase_overlap_score}). A likely explanation for this pattern is the distribution of essay lengths in the training set (Table~\ref{tab:trainingset}). Higher-scoring essays are substantially longer and more content-dense than lower-scoring essays, requiring more aggressive compression to satisfy downstream token limits. As the summarization procedure condenses these longer texts, information loss becomes more likely. This explains why semantic similarity, ROUGE scores, entity coverage, and keyphrase overlap all decrease as score level increases. Importantly, this indicates that summarization is not a neutral preprocessing step; it can systematically influence which information is preserved for scoring, particularly in high-quality essays that contain richer arguments and more nuanced structure.

Several limitations should be acknowledged. First, the study did not systematically examine prompt variation; only one summarization prompting strategy was used, leaving uncertainty about how alternative prompt designs might affect summary quality and downstream scoring. Second, only one embedding model was implemented, limiting generalizability across alternative semantic representation methods. Third, the choice of machine learning classifiers was guided by prior study results, meaning that other potentially strong downstream models were not fully explored. Fourth, the current experiments use a single public AES dataset, so the findings may not generalize to narrative, multilingual, discipline-specific, or high-stakes writing contexts without additional validation. Fifth, the reported results do not yet include direct comparisons with truncation, Longformer-style long-context encoders, direct zero-shot LLM scoring, open-source LLM summarizers, or recent SOTA AES systems such as NPCR, T-MES, CEAES, and GAT-AES. These omissions limit the strength of claims about relative performance.

Future research should therefore compare multiple prompting strategies, embedding models, classifier families, datasets, and baseline systems within a comprehensive ablation framework. A stronger follow-up benchmark should include stratified resampling or cross-validation, report confidence intervals or paired significance tests for QWK, and evaluate whether the observed score-band decline in summarization quality remains stable across datasets and writing genres. It would also be valuable to investigate score-aware or length-adaptive summarization methods that allocate different compression strategies to essays of varying lengths and quality levels. Such approaches may help reduce the performance drop observed for higher-scoring essays and improve the fairness, robustness, and reliability of automated essay scoring systems.

Beyond technical performance, this study highlights important social-scientific and educational implications of deploying automated essay scoring systems. In educational assessment contexts, AES systems are evaluated not only in terms of predictive accuracy but also with respect to fairness, accessibility, scalability, and alignment with pedagogical goals. A system that achieves strong performance at a lower computational cost can significantly expand access to high-quality writing assessment, particularly for institutions with limited technological and financial resources. This is especially relevant in large-scale educational settings—such as standardized testing programs, online learning platforms, and large university courses—where cost-efficient systems can enable faster feedback cycles, reduce grading workloads, and support more consistent evaluation practices. Importantly, faster turnaround times enable AES systems to be integrated into formative assessment practices, in which immediate feedback supports iterative revision, metacognitive reflection, and continuous skill development. In this sense, cost-efficient AI models do not merely reduce operational burden but also enhance the pedagogical value of writing assessment by enabling more frequent and responsive feedback loops between students and instructors.

At the same time, the finding that higher-quality essays are more difficult to summarize without information loss raises a critical concern regarding fairness and construct representation. Higher-scoring essays tend to exhibit greater argumentative complexity, richer evidence, and more sophisticated discourse structures, all of which are central to how writing proficiency is defined in educational contexts. If the summarization process disproportionately removes such features, AES systems may risk underrepresenting advanced writing ability and narrowing the construct being measured. This has direct implications for validity, as the system may inadvertently favor simpler, more easily compressible writing over more nuanced responses. From an instructional perspective, this could also shape student behavior, as learners may adapt their writing strategies to align with what the system captures most effectively. Therefore, preprocessing steps such as summarization should be carefully designed and evaluated across score levels to ensure that they preserve key dimensions of writing quality and do not introduce systematic bias against high-performing students.

More broadly, this study contributes to understanding automated essay scoring as a sociotechnical system embedded within educational practice rather than a purely technical prediction task. Model selection influences not only performance metrics but also resource allocation, environmental sustainability, and the feasibility of deployment across diverse educational contexts. More efficient models can reduce infrastructure demands and energy consumption, which is particularly important as AI-driven assessment systems scale globally. Furthermore, the hybrid framework proposed in this study—combining generative summarization, contextual embeddings, and handcrafted linguistic features—demonstrates how advanced neural methods can be integrated with interpretable indicators of writing. This integration can help educators better understand what aspects of writing are being evaluated, thereby supporting transparency, trust, and alignment with instructional objectives. Ultimately, effective AES systems must balance technical performance with educational responsibility, ensuring that they not only scale efficiently but also support meaningful, equitable, and pedagogically grounded assessment practices.

\section{Data Availability}
The data utilized in this research study is publicly available and can be found at \url{https://www.kaggle.com/competitions/learning-agency-lab-automated-essay-scoring-2/overview}. 

\section{Ethics Statement}
This study used only publicly available, de-identified secondary data from the Learning Agency Lab -- Automated Essay Scoring 2.0 competition. No new human participants were recruited, no intervention was conducted, and no personally identifiable student information was collected by the author. The study was therefore treated as not requiring additional ethics approval under the author's institutional practice for non-interventional public-data research.

\section{Declaration of Generative AI Use}
Generative AI tools were used as part of the research pipeline to summarize student essays, as described in the methodology. During manuscript preparation, generative AI assistance was also used for language editing, organization of revision responses, and clarity improvements. The author reviewed, edited, and approved all content and remains fully responsible for the accuracy, integrity, and interpretation of the manuscript.

\bibliographystyle{apacite}
\bibliography{bib}

\end{document}